%% file: main_arxiv.tex
\documentclass{article}
\usepackage{graphicx}
\usepackage{amssymb}

\usepackage[preprint]{corl_2026}

\usepackage{times}
\usepackage[numbers]{natbib}
\usepackage{multicol}
\input{preamble}

\usepackage{wrapfig}
\BeforeBeginEnvironment{wrapfigure}{\setlength{\intextsep}{0pt}}
\AfterEndEnvironment{wrapfigure}{\setlength{\intextsep}{0pt}}

\title{PanoVine: Whole-Body Visuomotor Control\\ for Soft Growing Vine Robot}

\author{
Yimeng Qin$^{*}$ \quad
Xiaomeng Xu$^{*}$ \quad
William Heap \quad
Aditi Oak \quad
Shuran Song$^{\dagger}$ \quad
Allison Okamura$^{\dagger}$ \quad \\\\
Stanford University \\\\
\large{\href{https://panovine-bot.github.io}{https://panovine-bot.github.io}}
\vspace{-8mm}
}

\begin{document}

\renewcommand{\thefootnote}{\fnsymbol{footnote}}
\footnotetext[1]{Equal Contributions} \footnotetext[2]{Equal Advising}

\maketitle

\begin{abstract}
Vine robots, a class of soft, growing robots, are suitable for navigating complex and confined environments due to their compliant bodies and self-supporting growth mechanism. However, hysteresis, tether interactions, and deformations make them difficult to predict and model, which in turn limits the effectiveness of conventional planning and control approaches. In this work, we present a data-driven, vision-based control framework for the first autonomous vine robot system. Our system integrates 19 cameras distributed along the robot's body to provide comprehensive feedback of both the robot state and the surrounding environment. Using this rich whole-body vision feedback, we train an end-to-end visuomotor policy from demonstrations for closed-loop autonomous control in complex environments. The policy efficiently aggregates information from distributed sensing while maintaining robustness to inaccurate robot states and actuation. Experimental results demonstrate that the learned policy enables robust navigation and manipulation in challenging scenarios, including steering through branched structures, climbing up slopes, traversing unsupported terrain, reaching objects precisely, and maneuvering through confined spaces and obstacles.
\end{abstract}


\input{text/intro}

\input{text/related}

\input{text/method}

\input{text/eval}


\section{Limitations and Future Work}
\vspace{-2mm}
The current system only employs RGB cameras operating at a small field of view (FOV), and lacks contact and force signals; higher-FOV cameras and distributed force and tactile sensing would provide more information on environmental context. 
Each task is learned from tens of teleoperated demonstrations on a single embodiment; training with more data would make the policy more generalizable and robust. Finally, the robot and sensing designs are based on engineering experience, automatically optimizing them based on task-specific requirements would enhance performance~\cite{xu2025dynamics, yi2025co}.

\section{Conclusion}
\vspace{-2mm}
We presented PanoVine, a vine robot platform and learning framework that brings whole-body vision and visuomotor policy learning to long, deformable soft growing robots. The system combines a 6\,m seven-DoF eversion-based body with 19 distributed cameras and a diffusion policy for whole-body visuomotor control. On real-world deployments the policy achieves 80\% success on a six-meter complex course chaining branch selection, slope climbing, unsupported-gap traversal, obstacle avoidance, and sharp turns, and 85\% success on precise object reaching, while single-camera and open-loop baselines fail completely. To our knowledge, this is the first system to demonstrate autonomous closed-loop control of a vine robot using only on-board sensing. We hope our system will encourage and facilitate future research on learning-based control for soft robots.

\acknowledgments{
The authors would like to thank the CHARM Lab and REALab members for their helpful discussions and feedback on the manuscript.
Xiaomeng Xu is supported by the Stanford Interdisciplinary Graduate Fellowship, and Yimeng Qin is supported by the Stanford Woods Institute for the Environment. This work was supported in part by NSF Awards \#2143601, \#2037101, and \#2132519, an Amazon Research Gift, Stanford System-X, the Stanford Woods Institute for the Environment, and the Stanford University Sustainability Accelerator. The views and conclusions contained herein are those of the authors and should not be interpreted as necessarily representing the official policies, either expressed or implied, of the sponsors.
}


\clearpage

\bibliography{reference}

\clearpage
\setcounter{section}{0}
\section*{Appendix}
\input{text/supp}

\end{document}

%% file: preamble.tex
\usepackage{graphics} 
\usepackage{epsfig} 
\usepackage{mathptmx} 
\usepackage{amsmath} 
\usepackage{amssymb}  
\usepackage{siunitx} 
\usepackage{textcomp}
\usepackage{gensymb} 
\usepackage{booktabs}
\usepackage{multirow}
\usepackage{xcolor}
\usepackage{algorithm}
\usepackage{algpseudocode}

\usepackage{enumitem}
\usepackage{xspace}
\usepackage{float}
\usepackage{comment}
\usepackage{mathtools}
\usepackage{soul}
\usepackage{natbib}
\usepackage[most]{tcolorbox}
\usepackage{xcolor}

\tcbset{
    greenbox/.style={
        colback=green!20, 
        colframe=green!60!black, 
        coltext=green!60!black, 
        boxrule=0.5mm,
        arc=2mm, 
        auto outer arc,
        boxsep=5pt,
        left=2mm,
        right=2mm,
        top=2mm,
        bottom=2mm,
    }
}

\usepackage{makecell}
\usepackage{rotating}
\usepackage[percent]{overpic}
\usepackage{contour}
\usepackage{courier}

\usepackage{subcaption} 
\usepackage[font=small]{caption}
\usepackage[percent]{overpic}

\usepackage[11pt]{moresize}

\usepackage{pifont}
\definecolor{MyDarkBlue}{rgb}{0,0.08,1}
\definecolor{airforceblue}{rgb}{0.36, 0.54, 0.66}
\definecolor{MyDarkGreen}{rgb}{0.02,0.6,0.02}
\definecolor{MyDarkRed}{rgb}{0.8,0.02,0.02}
\definecolor{MyDarkOrange}{rgb}{0.40,0.2,0.02}
\definecolor{MyPurple}{RGB}{111,0,255}
\definecolor{MyRed}{rgb}{1.0,0.0,0.0}
\definecolor{MyGold}{rgb}{0.75,0.6,0.12}
\definecolor{MyDarkgray}{rgb}{0.66, 0.66, 0.66}
\definecolor{MyPink}{rgb}{0.9, 0.33, 0.5}
\definecolor{MyCyan}{rgb}{0., 0.4, 0.4}

\definecolor{guidance_green}{RGB}{12,131,27}
\definecolor{theme_orange}{RGB}{255,138,0}
\definecolor{theme_blue}{RGB}{67,99,216}
\definecolor{theme_taro}{RGB}{219,176,234}

\definecolor{pure_green}{RGB}{0,255,0}
\definecolor{pure_red}{RGB}{255,0,0}

\makeatletter
\DeclareRobustCommand\onedot{\futurelet\@let@token\@onedot}
\def\@onedot{\ifx\@let@token.\else.\null\fi\xspace}

\makeatother

\usepackage[most]{tcolorbox}

\usepackage{xparse}
\usepackage{lipsum}

\def\exampletext{Example} 

\NewDocumentEnvironment{testexample}{ O{} }
{
    \colorlet{colexam}{theme_blue} 
    \newtcolorbox[use counter=testexample]{testexamplebox}{%
        empty,
        title={\exampletext: #1},
        attach boxed title to top left,
        minipage boxed title,
        boxed title style={empty,size=minimal,toprule=0pt,top=2pt,left=2mm,overlay={}},
        coltitle=colexam,fonttitle=\bfseries,
        before=\par\nonskip\noindent,parbox=false,boxsep=0pt,left=2mm,right=0mm,top=2pt,breakable,pad at break=0mm,
        before upper=\csname @totalleftmargin\endcsname0pt, 
        after=\par\nonskip,
        overlay unbroken={\draw[colexam,line width=.5pt] ([xshift=-0pt]title.north west) -- ([xshift=-0pt]frame.south west); },
        overlay first={\draw[colexam,line width=.5pt] ([xshift=-0pt]title.north west) -- ([xshift=-0pt]frame.south west); },
        overlay middle={\draw[colexam,line width=.5pt] ([xshift=-0pt]frame.north west) -- ([xshift=-0pt]frame.south west); },
        overlay last={\draw[colexam,line width=.5pt] ([xshift=-0pt]frame.north west) -- ([xshift=-0pt]frame.south west); },%
    }
    \begin{testexamplebox}}
        {\end{testexamplebox}\endlist}

\AfterEndEnvironment{testexample}{\color{black}}

\DeclareMathAlphabet{\mathcal}{OMS}{cmsy}{m}{n}

%% file: text/intro.tex
\begin{figure}[ht]
  \centering
  \includegraphics[width=\linewidth]{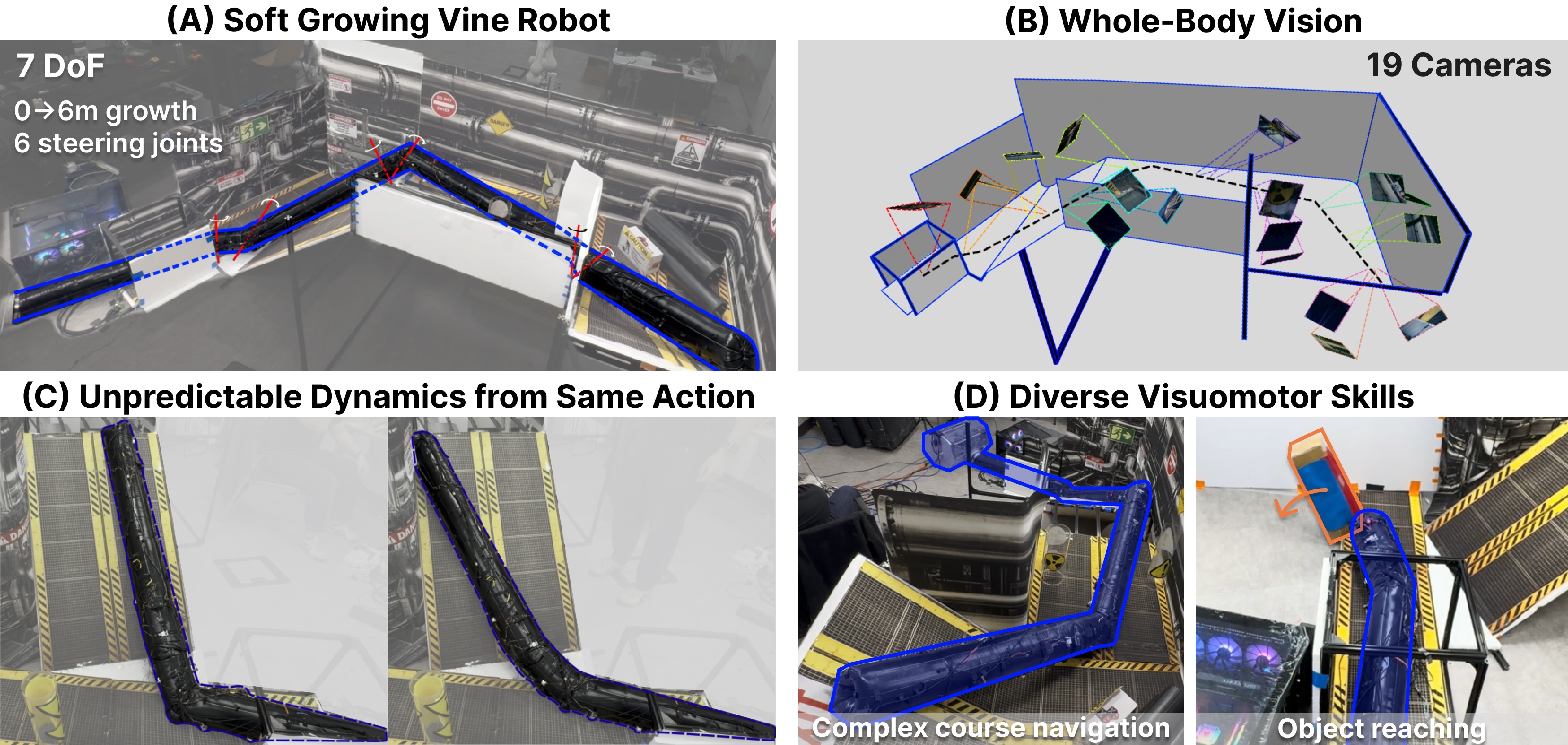}
  \vspace{-0.5cm}
  \caption{
      \textbf{PanoVine System} features (A) a \SI{6}{m} soft growing vine robot with (B) 19 cameras distributed along the robot's body. 
      (C) The system is challenging to control due to its unpredictable robot dynamics, where same action command can lead to drastically different robot configurations due to unpredictable buckling locations, soft-material hysteresis, and interactions with the environment. 
      (D) 
      The PanoVine System addresses the challenges and enables diverse navigation and manipulation skills by learning a whole-body visuomotor policy that leverages whole-body vision feedback. 
      %
}
  \vspace{-0.5cm}
  \label{fig:overview}
\end{figure}

\section{Introduction}
Maintenance and inspection of industrial assets such as pipes, pressure vessels, and underground infrastructure remain challenging due to their confined geometry, limited communication access, and long deployment distances. Narrow passages, sharp bends, branches, and unsupported structures make these environments difficult to navigate for rigid robots that rely on wall-contact locomotion~\cite{Yoon-GuKim2011,Ren2019}; limited communication access also makes drones that rely on GPS signals infeasible~\cite{GPS}.

Vine robots, a class of pneumatic soft growing continuum robots, are well suited to these scenarios~\cite{vine, blumenschein2020design}. They \textbf{grow} in length and extend at the tip by using internal fluid pressure to evert body material supplied from a fixed base. This growth mechanism allows vine robots to extend along tortuous paths to significant lengths, generate large propulsion forces while experiencing negligible sliding friction w.r.t. the environment, squeeze through apertures smaller than their body diameter, and have a lightweight self-supporting body capable of carrying sensors and trailing payloads.

The vine robot we employ can actively control the angle of multiple joints distributed throughout its body to \textbf{steer} towards desired directions or avoid obstacles, select branches, climb slopes, and pass through gaps, providing the dexterity needed to navigate confined, unstructured environments.

However, the same properties that make vine robots attractive also make it difficult to control their complex growing and steering motions for intricate navigation and manipulation:
\vspace{-0.15cm}
\begin{itemize}[nosep,leftmargin=3mm]
    \item \textbf{Unpredictable robot dynamics.} The robot behavior is shaped by material compliance, hysteresis, buckling, body-environment interaction, and tension in the un-everted tail, all of which are difficult to predict and model. As shown in Fig.~\ref{fig:overview}~C, open-loop replay of the exact action trajectory can lead to significantly different robot configurations across different trials~\cite{blumenschein2020design}.
    \item \textbf{Limited onboard sensing.} The robot undergoes intricate whole-body motions and deformations and spans a large and complex workspace (Fig.~\ref{fig:overview}~A), making limited onboard sensing (e.g., IMU, single camera) insufficient for reliable control~\cite{qinpipes, 11020809}.
\end{itemize}


To tackle these challenges, in this work, we propose 
\textit{a whole-body vision and visuomotor policy learning system for controlling vine robots in complex environments.}
To our knowledge, this is the first vine robot system capable of autonomous closed-loop control using on-board sensing alone.

To address unpredictable robot dynamics, we learn a visuomotor policy from demonstrations. High-dimensional visual feedback directly captures body deformation, contact state, and surrounding geometry, allowing a \textbf{whole-body visuomotor policy} trained end-to-end to compensate for uncertainty in robot states and dynamics. The policy is specifically designed to handle three challenges: aggregating information from distributed sensing, maintaining robustness to unreliable state estimation and actuation, and learning long-horizon behaviors where sparse steering actions are critical.

Such a policy requires rich sensory coverage across the entire robot body. Because the robot undergoes intricate whole-body motions over large workspaces, a single camera is insufficient to capture the observations a whole-body policy needs. We therefore equip the robot with a \textbf{whole-body vision} suite of 19 cameras attached along its body. The cameras are gradually revealed and exposed to the environment as the robot grows during eversion, and they collectively provide comprehensive feedback of both the robot and its environment (Fig.~\ref{fig:overview}~B). When the robot is short, few cameras are revealed; more are revealed as it lengthens and gains degrees of freedom.

In summary, our contributions are:
\vspace{-0.15cm}
\begin{itemize}[nosep,leftmargin=3mm]
\item A vine robot platform with distributed whole-body vision to provide multi-perspective feedback of both the robot state and its environment.
\item A learning-based visuomotor control framework for vine robots with design choices that improve robustness to perception and actuation uncertainty.
\item Demonstration of challenging long-horizon navigation and manipulation tasks, including steering through branched structures, traversing unsupported terrain, maneuvering through cluttered obstacles, and reaching objects in the environment. Result videos are best viewed on \href{https://panovine-bot.github.io}{https://panovine-bot.github.io}.
\end{itemize}

%% file: text/related.tex
\section{Related Work}

\textbf{Control and Planning for Soft Growing Robots.}
Recent research in control and planning for soft growing robots, particularly vine robots, can be broadly categorized into model-based and data-driven approaches. Model-based methods typically rely on kinematic models, such as the piecewise constant curvature assumption~\cite{webster2010constant, Blumenschein2017ModelingOB, Blumenschein2020GeometricSF, Ataka2020}, to plan tip trajectories by controlling the robot's curvature. For instance, \citeauthor{greer_tip_extension_2019} combined a kinematic model with visual servoing to regulate steering based on image-space error, and \citeauthor{Watson} introduced a Jacobian-based adaptive control framework to enable deployment in contact-rich environments. Other works exploit obstacle contact for navigation~\cite{greer2020, Selvaggio2020}. However, these methods often depend on prior knowledge of obstacle geometry or onboard sensing for real-time SLAM, which is difficult to provide in confined, cluttered environments — particularly on everting robots, where tip-mounted sensors add mechanical complexity and compromise reliability. The nonlinear, hysteretic behavior of soft materials further frustrates physics-based modeling: buckling and wrinkling during growth produce discontinuous deformations~\cite{Blumenschein2017ModelingOB}, and as the robot lengthens, accumulated tail tension can cause the body to buckle or deviate from the intended path in unsupported sections~\cite{Coad2020}. In contrast, PanoVine learns an end-to-end visuomotor policy from distributed whole-body vision that directly observes contact, deformation, and surrounding geometry, bypassing both the obstacle-geometry and body-modeling assumptions that limit model-based pipelines on long, tortuous deployments.


\textbf{Learning-Based Methods for Soft Robots.}
To address the modeling limitations inherent in soft robotic systems, recent works have explored learning-based control framework. \citeauthor{Hussieny2024} developed a deep reinforcement learning framework for obstacle-aware navigation, but the policy was only demonstrated in simulation. 
\citeauthor{Kalibala2025} proposed a learning-based model predictive control framework trained on simulated trajectories, but this was also unverified in the real world. \citeauthor{jitosho2023reinforcement} demonstrated reinforcement learning–based dynamic control on short vine-robot–like soft arms, but their reliance on finite-element discretization limits scalability to long, continuously growing robots. \citeauthor{Haggerty2023} use a Koopman-based approach to learn an explicit dynamical model that enables model-based control; however, it relies on carefully chosen state representations and does not scale well to long, multi-segment systems.
In contrast, we learn an end-to-end visuomotor policy directly from demonstrations, which bypasses the challenge of modeling system dynamics and planning high-DoF actions.

\textbf{Whole-Body Sensing for Robotics.}
Prior work in whole-body sensing investigates distributed range, force, tactile, and vision sensing to enhance robot perception, addressing challenges in collision avoidance, contact detection, compliant motion, and whole-body manipulation. Range sensing enables semi-autonomous navigation for snake robots~\cite{tanaka2015range} and safe human–robot interaction~\cite{qi2022safe}. Distributed force-torque sensors can enable compliant motions~\cite{kollmitz2018whole, goncalves2022punyo, murooka2025tact, choi2026wild, xu2026compliant}, while tactile sensing provides dense contact feedback~\cite{dahiya2009tactile,liu2023enhancing,dean2019whole,jiang2024hierarchical}. While effective for detecting nearby obstacles and physical contacts, these sensing modalities provide limited semantic understanding of the environment and task context. Vision offers richer geometric and semantic information beyond immediate contact. Recent works~\cite{xu2026hommi, punamiya2026egoverse, xiong2025vision} employ egocentric and wrist cameras for bimanual mobile robots, but still suffer from occlusions and are limited to end-effector-only manipulation. RoboPanoptes~\cite{XuX-RSS-25} explores whole-body vision on a rigid serial robot by leveraging distributed visual sensing for whole-body manipulation. Our work applies the concept of whole-body vision to a soft growing vine robot, which enables continuous, multi-perspective observation of both the robot body and the surrounding environment throughout growth and steering.

%% file: text/method.tex
\begin{figure}[t]
  \centering
  \includegraphics[width=0.95\columnwidth]{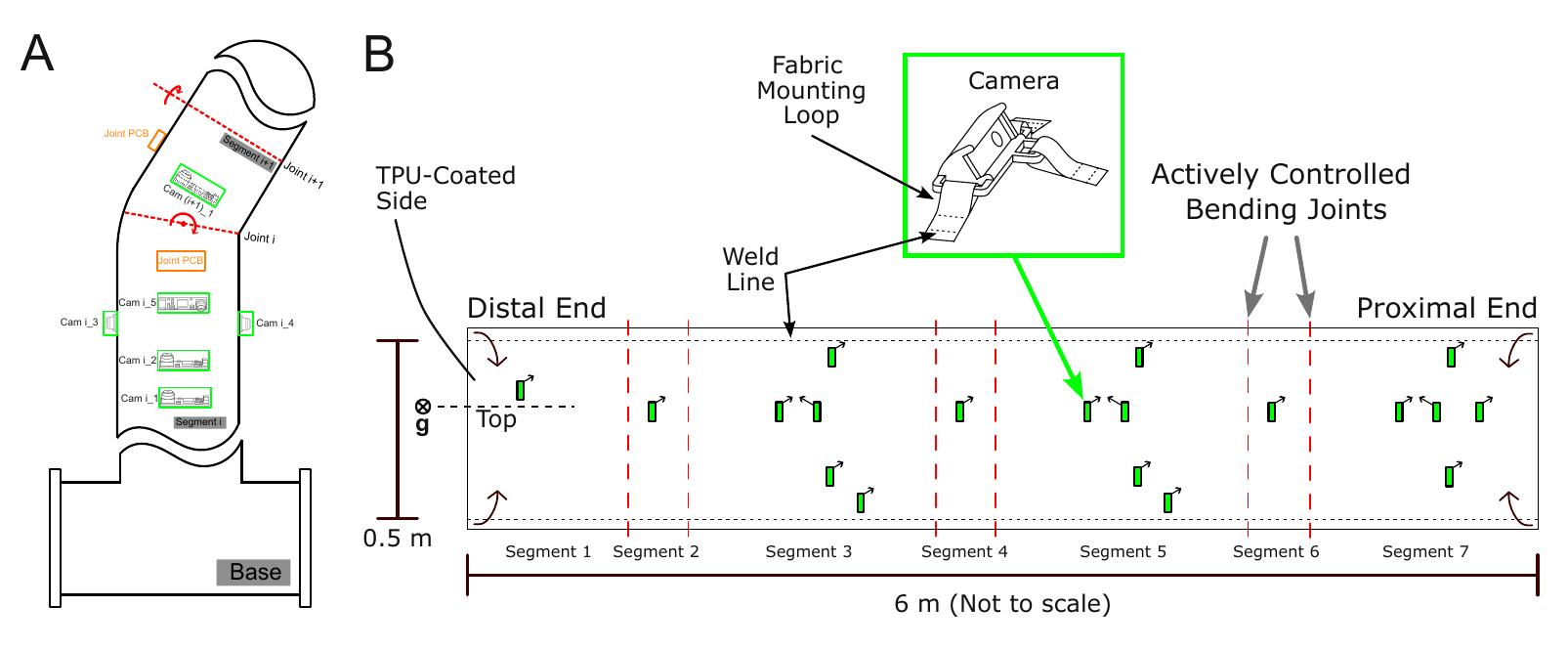}
  \vspace{-4mm}
  \caption{\textbf{PanoVine Robot Design.} (A) Scalable design of a single segment, showing the locations of the cameras and joints. (B) Placement of the 6 revolute joints and 19 RGB cameras distributed across the 7 segments of a 6 m long, 0.5 m diameter robot. Each joint is actively controlled to bend, changing the relative angle of adjacent segments. Cameras are attached to the TPU-coated side via welded fabric mounting loops.
}
  \vspace{-8mm}
  \label{fig:system}
\end{figure}
\section{Vine Robot System}
\subsection{Robot Design and Fabrication}

\textbf{Robot Design.} The robot has \textit{seven controllable degrees of freedom}, one from extension at the tip, and six from revolute joints.
For the \textit{tip extension}, similar to prior vine robot designs, the body of the vine robot used in this work consists of an airtight fabric tube flexible enough to be inverted into itself. When the inside of the tube is pressurized, the inverted material is everted at the tip of the robot body, and the robot body lengthens. We controlled the lengthening rate by spooling the inverted robot body material on motorized drum. For this work, we built a 6 m long cylindrical body 0.16 m in diameter. To enable robot \textit{re-configuration and steering}, we attached 3 pairs of approximately revolute joints along the body of the robot. The joints within each pair are orthogonal to each other and the robot body's central axis, and parallel to a joint in the other pairs. Each joint is actively controlled in bending. Orthogonal joints within a pair are spaced 0.15 m apart, and each pair of joints is spaced 1 m apart.
The design allows for high scalability and customizability for different environments by fabricating a longer body tube or adding more revolute joints.
 
\textbf{Robot Fabrication.} The primary material is 70-denier nylon ripstop fabric with one side coated with thermoplastic polyurethane (TPU). The vine robot body was fabricated by using an ultrasonic welder (Vetron 5064) to weld fabric attachment features for the actuators, local computing units, cameras, and cables to a fabric sheet. 

\textbf{Camera Placement.} The motors, computing units, and cameras are housed in 3D-printed cases with tabs, and secured to the vine robot body with fabric loops. Nineteen cameras are mounted to the robot body at a 60 degree angle relative to the axis of the robot body (Fig.~\ref{fig:system}), facing either the distal or proximal end of the robot to ensure that a wide range of the environment is observed. One camera is mounted between the proximal end of the vine robot body and the first joint, and six are mounted after every joint pair. Six computing units are mounted before and after each joint pair.

\begin{wrapfigure}{r}{.65\linewidth}
  \centering
  \vspace{-6mm}
  \includegraphics[width=\linewidth]{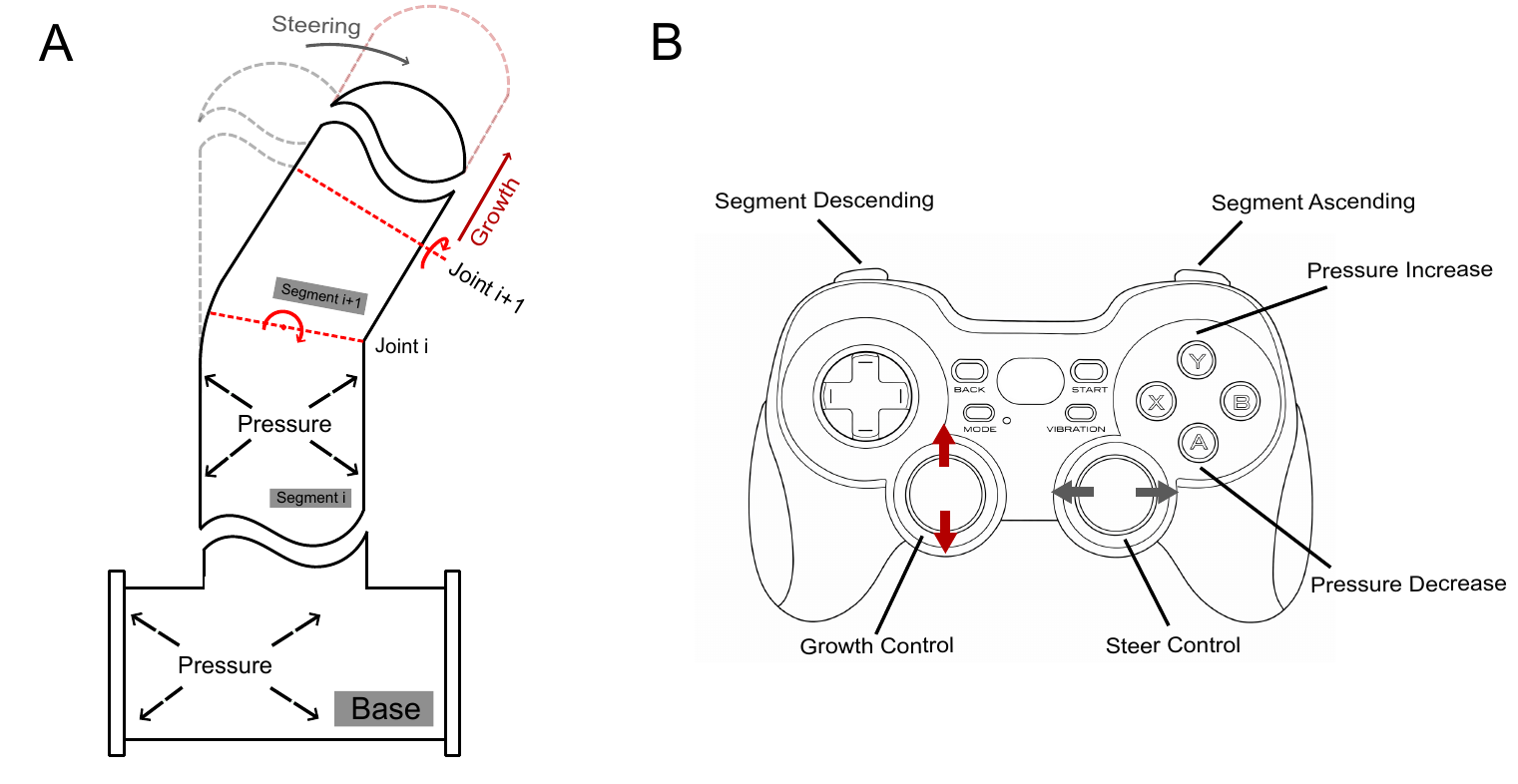}
  \vspace{-6mm}
  \caption{\textbf{Data Collection Interface.} A joystick controls the robot by independently commanding joint-space steering and axial growth.
  }
  \vspace{-2mm}
  \label{fig:teleop}
\end{wrapfigure}

\subsection{Data Collection Interface}
As shown in Fig.~\ref{fig:teleop}A, the operator teleoperates the multi-segment vine robot using a joystick (Logitech G F710). The active robot segment is switched using the RT and LT buttons. Joint motion of the selected segment is controlled via rate control on the right joystick, while the growth velocity is mapped to the left joystick. The robot body pressure, and thus stiffness, is adjusted using the X and A buttons to decrease and increase the value, respectively. 
%
%
Joystick commands are communicated through the Robot Operating System (ROS) at \SI{30}{Hz}; joint angle and IMU data over RS-485 at \SI{10}{Hz}; base-station motor encoder and torque measurements via USB at \SI{50}{Hz}; and camera data via USB at 19.5~fps. We pre-characterize the relationships between joint angles and joint sensor readings, as well as between the robot odometer and deployed robot length, using calibrated mapping functions. 


\section{Whole-Body Visuomotor Policy}





\begin{wrapfigure}{r}{0.6\linewidth}
\centering
\includegraphics[width=\linewidth]{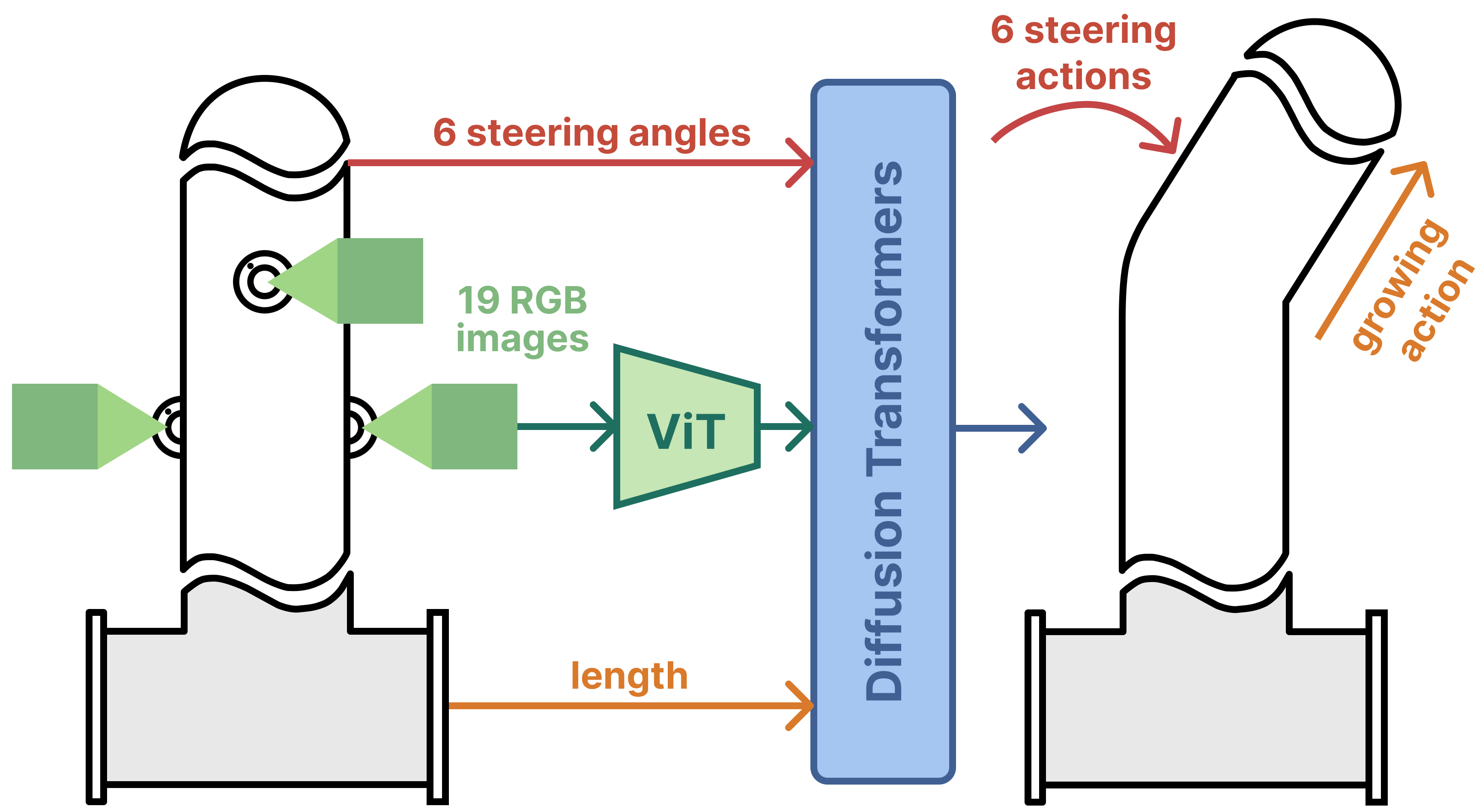}
\caption{\textbf{PanoVine Whole-Body Visuomotor Policy.}
The environment and robot states are observed through PanoVine's 19 cameras and growing and steering sensors. The 19 RGB images are represented by the class token of a vision foundation model. The vision tokens along with proprioception are taken by a diffusion transformers policy to predict growing and steering actions.
}
\vspace{-0.2cm}
\label{fig:policy}
\end{wrapfigure}

We learn a whole-body visuomotor policy from teleoperated demonstrations, as shown in Fig.~\ref{fig:policy}. At each control step, the policy maps a history of sensory observations to an action chunk that specifies how the robot should grow and steer over the next several steps. 

The policy receives observations $o=(\mathcal{I},q)$, where $\mathcal{I}$ denotes the RGB streams from the 19 body-mounted cameras, and $q$ denotes proprioception, comprising the joint angles and a base length signal derived from the growth spool encoder. The policy predicts an action chunk $a\in\mathbb{R}^{\tau\times d}$ consisting of the command base length and joint angles over a horizon $\tau$ ($\tau=8$ in our experiments). Both observations and actions are sampled at \SI{5}{Hz}. We adopt the diffusion transformers policy~\cite{chi2023diffusion, XuX-RSS-25} as our backbone, and introduce three
design choices that target the specific control challenges of a long, deformable, whole-body-sensed vine robot: (i)~aggregating distributed vision efficiently, (ii)~remaining robust to unreliable proprioception and actuation, and (iii)~preventing sparse but decisive steering events from being drowned out by long stretches of pure growth.

%

\subsection{Learning Multi-View Correspondences with a Cross-Attention Transformers}
Each body-mounted camera observes only a fragment of the full robot-and-environment state, and their relative poses change continuously as the body grows, buckles, and deforms under contact. Explicit extrinsic calibration is unreliable in this regime, so we instead let the network learn implicit cross-view correspondences.
Specifically, we apply color jittering and random dropouts~\cite{XuX-RSS-25} on the images (each resized to $224\times224\times3$) and feed them into a pretrained CLIP Vision Transformers model~\cite{dosovitskiy2020image, radford2021learning, xu2023jacobinerf}, where we take the predicted class token for each image as a compact per-view embedding. 
Low-dimensional proprioceptive streams (length and steering angles) are concatenated with the vision tokens, and the diffusion transformers decoder cross-attends actions to this observation.
This cross-attention learns task-specific semantic correspondences between multi-modal observations (multi-view images and proprioception) and robot actions.

\subsection{Relative Proprioception and Action Representation for Robust Control}
We express proprioception and the action chunk \emph{relative to the latest observation frame} in the window: each channel is subtracted from its value at the current observation time.
We also retain \emph{absolute} channels for length and joints so the policy still receives unambiguous global context as inputs. At inference time, we subtract the most recent proprioceptive reference from the observation buffer before inference and convert predicted relative actions back to absolute commands by adding the same reference.
This relative action representation ensures smooth transitions between action chunks and increases robustness to uncertainty in the robot's absolute state under actuation uncertainties and hysteresis.

\subsection{Rebalancing Steering and Growing Windows for Reactive Control}
\label{sec:rebalance}
Teleoperated trajectories are dominated by long segments of nearly pure growth with zero joint angle changes, while steering motion is comparatively sparse.
We label each training sequence window according to whether joint angles change over the action horizon relative to a threshold.
Then we rebalance the training dataset towards the same ratio of steering windows and pure growth windows.
This ensures the policy can learn critical and reactive steering actions for turning, climbing, bending etc. and avoid overfitting to growing actions.

%% file: text/eval.tex
\section{Evaluation}
We evaluate PanoVine on two challenging real-world tasks that stress different aspects of whole-body vision and visuomotor policy: long-horizon navigation through a complex course and precise reaching toward objects placed at unknown locations.

\subsection{Complex Course Navigation}

\begin{figure}[t]
    \centering
    \includegraphics[width=0.95\linewidth]{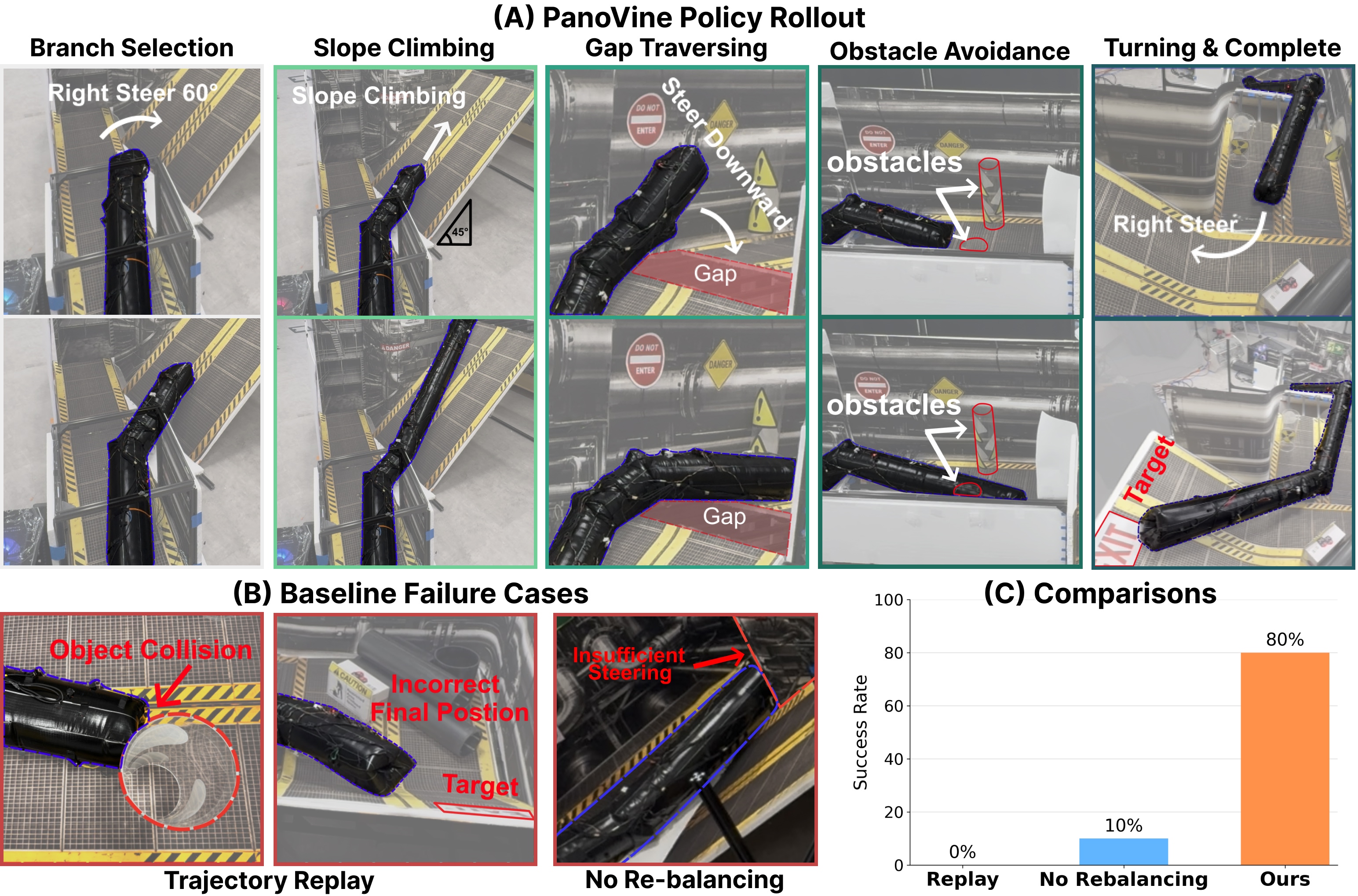}
    \vspace{-2mm}
    \caption{\textbf{Complex Course Navigation.} (A) Autonomous policy rollout of PanoVine, demonstrating long-horizon navigation skills in a complex environment, including steering through branched structures, climbing slopes, traversing unsupported gaps, avoiding obstacles, and making sharp turns. (B) Typical baseline failure cases. \texttt{Trajectory Replay} often collides with obstacles and fails to reach the correct final position. The \texttt{No Re-balancing} baseline undergoes insufficient steering. (C) Quantitative comparisons.
    }
    \vspace{-6mm}
    \label{fig:3d_course}
\end{figure}

\textbf{Task:}
As shown in Fig~\ref{fig:3d_course}~(A), the robot needs to navigate through a complex course that spans \SI{6}{m} long and \SI{1.5}{m} tall. First, after growing for \SI{0.6}{m}, the robot steers to the right for approximately \SI{90}{\degree} to enter a branch structure. Then, the robot climbs up a \SI{45}{\degree} slope, which is followed by an unsupported gap structure at the end, requiring it to lift up and then bend down to traverse the gap. Next, based on the placements of the obstacles in the environment, the robot needs to steer at precise angles to the right or left to pass through the obstacles without colliding. Finally, the robot turns at the correct moment at a narrow, sharp turn and climbs up another slope to reach the final course exit.

\textbf{Capability:}
\textit{Long-horizon control:} The course chains together five different skills---branch selection, slope climbing, unsupported gap traversal, obstacle avoidance, and sharp-turn maneuvering---over long continuous growth.
Small steering errors early could compound and affect downstream subtasks, so the policy must sustain accurate whole-body actions across the entire trajectory.

\begin{wrapfigure}{r}{0.6\linewidth}
\centering
\includegraphics[width=\linewidth]{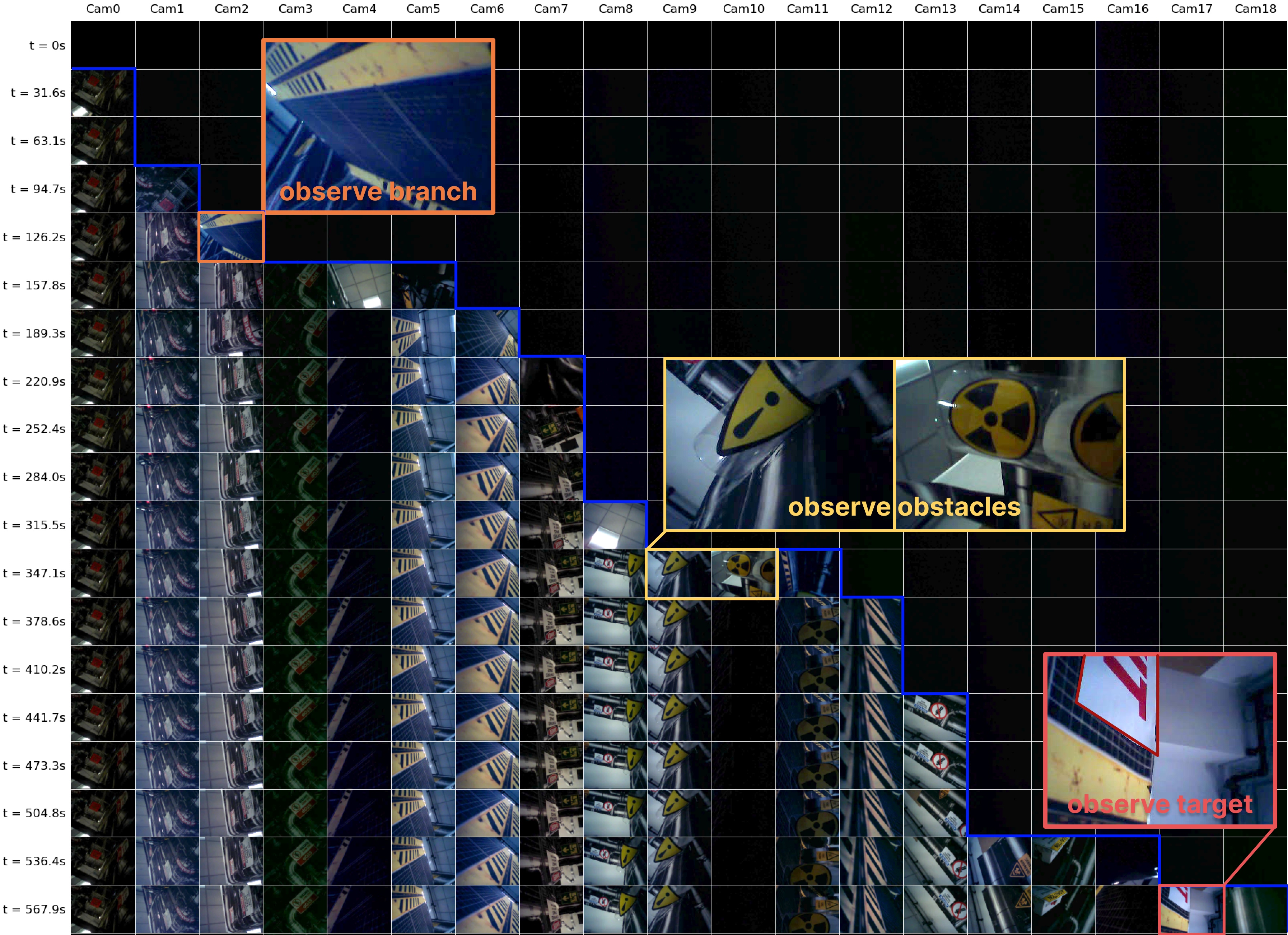}
\vspace{-6mm}
\caption{\textbf{Camera Views throughout Course Navigation.}
}
\label{fig:views}
\end{wrapfigure}

\textit{Reactive steering from visual feedback:} The robot's configuration at any given moment is only loosely predictable from the action history due to actuator force, material compliance, hysteresis, base buckling, and complex environmental interactions. The policy must therefore close the loop on its own body state at every step, using distributed visual feedback for reactive correction rather than committing to a precomputed trajectory. For example, in Fig.~\ref{fig:views}, the robot steers right when observing the branched structure, steers to avoid obstacles when they come into view, and stops growing after it observes the exit sign.

\textbf{Data Collection:} We collected 41 demonstrations with randomized obstacle placements and robot initial configurations (base buckling and initial length, initial orientation, and state of the joint actuators) with each demonstration taking an average duration of 12 minutes.

\textbf{Test Scenarios:} We ran 10 rollouts with randomized obstacle locations and robot initial configurations for all methods.

\textbf{Performance:}
\texttt{Ours} whole-body visuomotor policy achieves an 80\% success rate on the full course, demonstrating precise and reactive control from visual feedback that is robust to uncertainty in robot states and dynamics.
Occasional failure cases include getting stuck in the unsupported gap when the bending angle is insufficient to clear it, and colliding with obstacles in the final narrow corner when the last turn is initiated with an insufficient turning angle.

\texttt{Trajectory Replay} baseline, where we replay the recorded action trajectory of a successful teleoperated demonstration open-loop on the same course, fails on every trial (0\% success rate). Typical failure cases (Fig.~\ref{fig:3d_course}~(B)) include colliding with obstacles whose positions differ from the demonstration, and falling short of the goal because base buckling reduces the deployed length below what the replayed growth commands assume.
This confirms that the course is not solvable open-loop and that closed-loop visual feedback is required throughout the task.

\texttt{No Re-balancing} baseline, where we train the whole-body visuomotor policy on raw demonstrations without rebalancing steering and growing windows (Sec.~\ref{sec:rebalance}), achieves only 10\% success rate. As shown in Fig.~\ref{fig:3d_course}~(B), it fails at the first branch selection phase with insufficient steering, demonstrating that our rebalancing strategy is critical for reactive and accurate steering control.

\subsection{Object Reaching}
\begin{figure}[t]
    \centering
    \includegraphics[width=\linewidth]{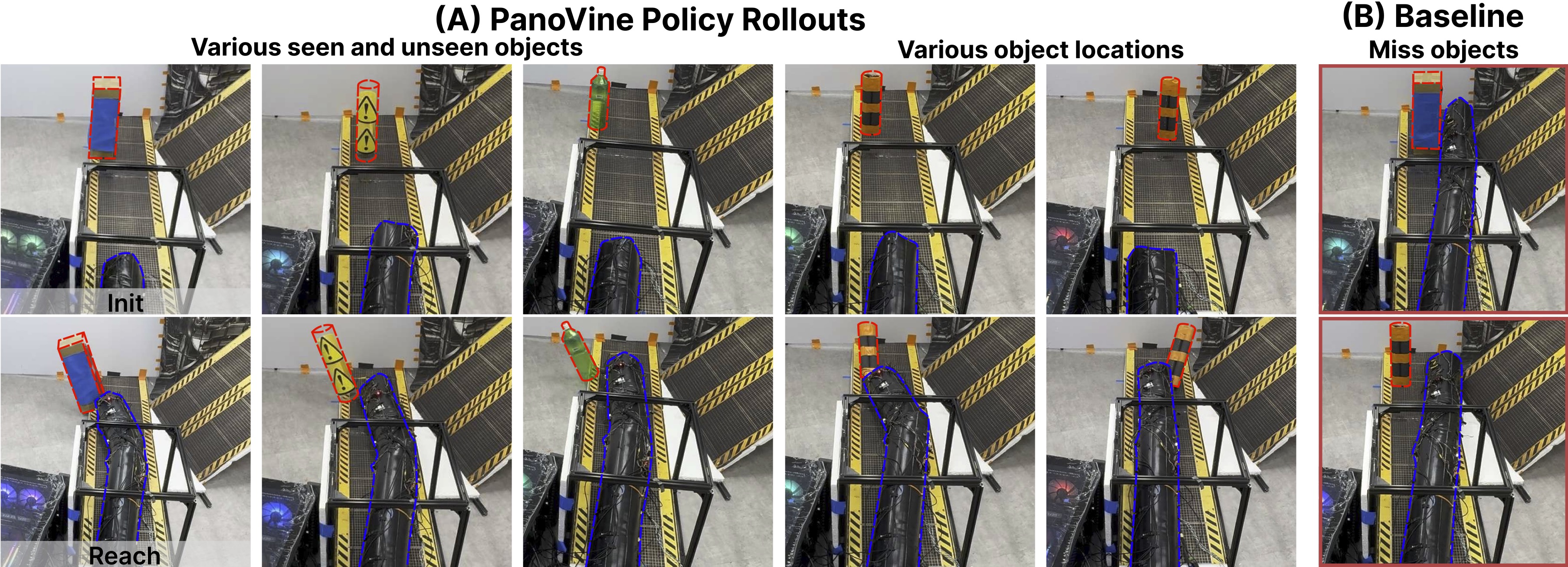}
    \vspace{-6mm}
    \caption{\textbf{Object Reaching.} (A) Autonomous PanoVine policy rollouts, demonstrating precise reaching of various objects placed at different locations, achieving \textbf{85\%} success rate. (B) \texttt{Single Camera Policy} baseline always misses objects, resulting in a \textbf{0\%} success rate.
    }
    \vspace{-6mm}
    \label{fig:object_reach}
\end{figure}

\textbf{Task:} As shown in Fig.~\ref{fig:object_reach}~(A), the robot needs to reach and touch objects \SI{2}{m} away at different locations w.r.t. the robot. The robot first extends forward for a distance, then when seeing the object in view, it steers left or right towards the object to reach it precisely and knock it down.

\textbf{Capability:}
\textit{Precise steering:} Successful contact requires aligning the tip with the object to within a small angular tolerance after \SI{2}{m} of growth, where even small joint-angle errors translate into large lateral offsets at the tip.

\begin{wrapfigure}{r}{0.35\linewidth}
\centering
\vspace{-2mm}
\includegraphics[width=\linewidth]{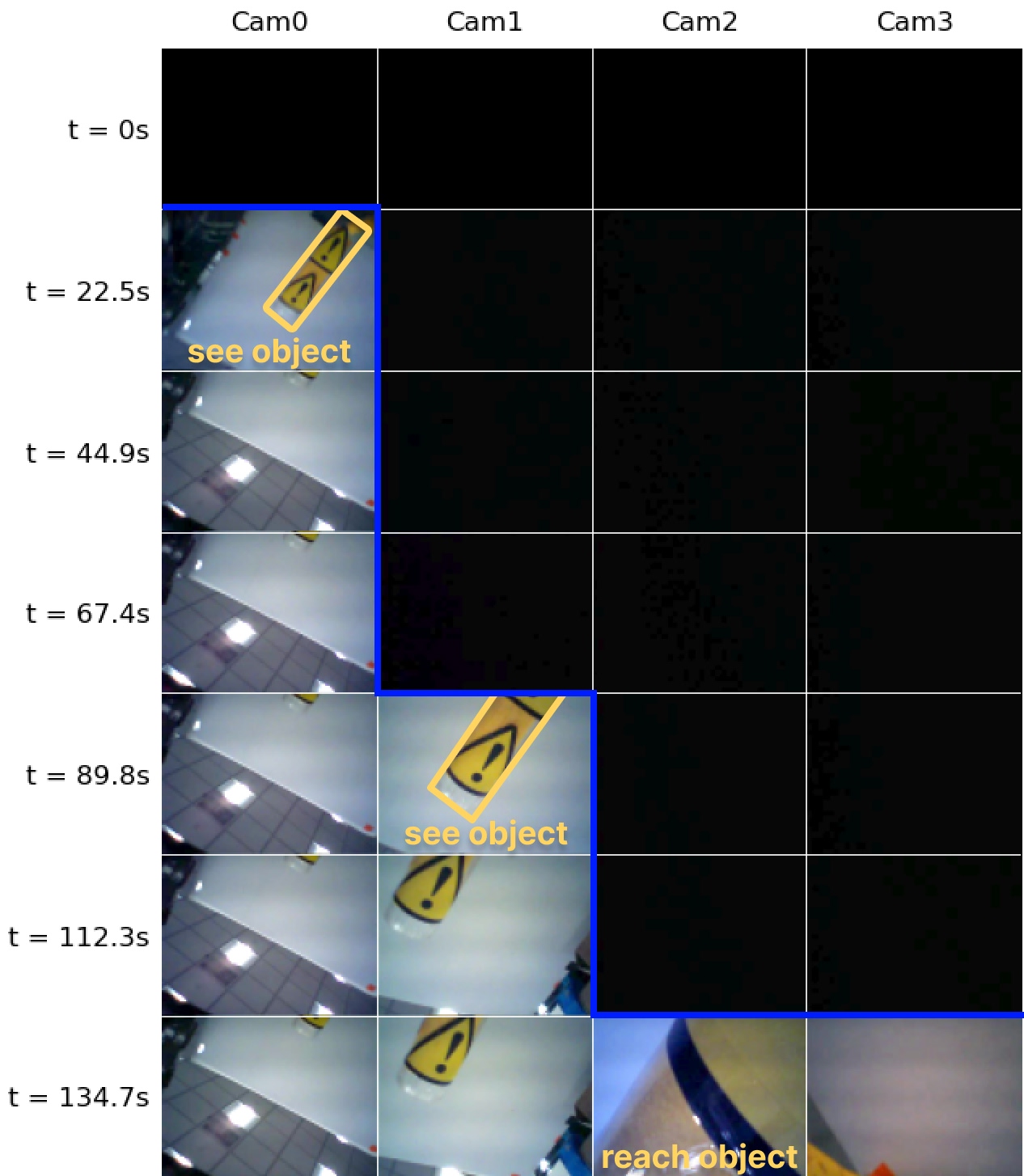}
\vspace{-6mm}
\caption{\textbf{Camera Views throughout Object Reaching.}
}
\vspace{-6mm}
\label{fig:views_object}
\end{wrapfigure}

\textit{Generalization across objects and placements:} Test objects include an object not seen during training and five distinct starting locations, so the policy cannot memorize a fixed trajectory. It must instead ground the steering command on the current visual appearance of the object across multiple cameras (Fig.~\ref{fig:views_object}), regardless of which specific object is present or where it is placed in the workspace.

\textbf{Data Collection:} We collected 80 demos on 4 objects with randomized locations. Average demo duration is 3 minutes.

\textbf{Test Scenarios:} We ran 20 rollouts on 4 objects (three seen and one unseen) starting from 5 different locations, and using the exact same test cases for all methods.

\textbf{Performance:}
\texttt{Ours} achieves an 85\% success rate, demonstrating precise visually-grounded steering under uncertainty in both target location and object geometry and appearance. 
The policy reactively adjusts its bending angle incrementally as the object becomes visible to successively more body cameras (Fig.~\ref{fig:views_object}), indicating that the policy successfully learned multi-view correspondences. 

\texttt{Single Camera Policy} baseline, where we only input the first camera's image into the policy, achieves 0\% success rate. The base camera cannot observe the object when the robot extends out and occludes it, or when the robot steers and the object goes out of view. The policy fails to turn and adjust towards the correct direction; thus, it always misses the object and grows past it (Fig.~\ref{fig:object_reach}~(B)).

%% file: text/supp.tex
\section{Sensing and Electronics}
The sensing and electronics system is divided into on-body components and base components. The on-body components comprise 19 USB webcams (GC0307 sensors), 12 magnetic rotary encoders (Pololu), and six custom local computing units. Each local computing unit integrates a Teensy~4.0 microcontroller (MCU), dual motor drivers (DRV8838), and an onboard power management system. 

At the base, a pressure regulator (QB3, Proportion-Air) is used to control the internal body pressure, and a high-torque brushless DC motor (CubeMars AK80-9) drives a motorized spool to regulate the robot growth velocity. A PCIe-to-USB interface board supports the high-bandwidth USB camera system, while a lead MCU mediates communication between the base computer and the distributed local computing units.
Full-duplex RS-485 communication is employed to enable bidirectional data exchange. In the forward direction, the lead MCU transmits motor control commands to individual local MCUs, specifying pulse-width modulation (PWM) duty cycle and pulse duration. In the reverse direction, each local MCU acquires motor encoder measurements and IMU data via analog and I\textsuperscript{2}C interfaces, respectively, and transmits the sensor data back to the base. To prevent data collisions on the shared RS-485 bus, a circular token-passing protocol is implemented, allowing only one local MCU to transmit at any given time.
All USB cameras stream visual data via USB~2.0 at a resolution of 320~×~240 pixels and 30~fps, with the base computer interfacing through the PCIe expansion board. Notably, the maximum robot length of 4.2~m remains within the practical USB~2.0 communication distance ($<5$~m), ensuring reliable data transmission.

\section{Policy Training Details}
\textbf{Observations and actions.}
We use a short observation history of $T_o{=}2$ steps and predict an action horizon of $T_p{=}8$ steps at \SI{5}{Hz} (downsampled by a factor of 3 from \SI{15}{Hz} demonstrations). Observations include 19 RGB camera streams ($3{\times}224{\times}224$ each) and proprioception. Proprioception comprises the robot's 6 segment joint angles and the base extension length, each provided in both an absolute and a latest-frame-relative form. Actions are 7-dimensional, consisting of the 1-dimensional base extension length and the 6 segment joint angles ($7=1+6$), predicted in the relative representation.

\textbf{Model.}
We use Diffusion Policy with a Diffusion Transformer backbone~\cite{chi2023diffusion}. The diffusion model is conditioned on global observation embeddings and predicts noise ($\epsilon$-prediction) with a DDIM scheduler~\cite{song2020denoising} using a squared-cosine $\beta$ schedule. We use $50$ training timesteps, $16$ inference steps, and input perturbation $0.1$. The
DiT uses embedding dimension $768$, depth $7$, $8$ heads, and attention dropout $0.1$. The observation encoder finetunes a CLIP-pretrained \texttt{ViT-B/16} backbone~\cite{dosovitskiy2020image}~(\texttt{vit\_base\_patch16\_clip\_224.openai}) shared across all camera streams, aggregating the \texttt{CLS} token, with ColorJitter augmentation. We maintain an exponential moving average (EMA) of the weights (power $0.75$, max decay $0.9999$).

\textbf{Optimization.}
We use AdamW~\cite{loshchilov2017decoupled} with a cosine learning rate schedule (2000 warmup steps) starting at a learning rate of $3{\times}10^{-4}$ for the diffusion model and $3{\times}10^{-5}$ for finetuning the vision backbone, weight decay $1{\times}10^{-6}$, and betas $(0.95, 0.999)$. We use a batch size of 32 and train our policy and all baselines for 500 epochs.

\section{Additional Camera Visualization}
\begin{figure}[t]
  \centering
  \includegraphics[width=\columnwidth]{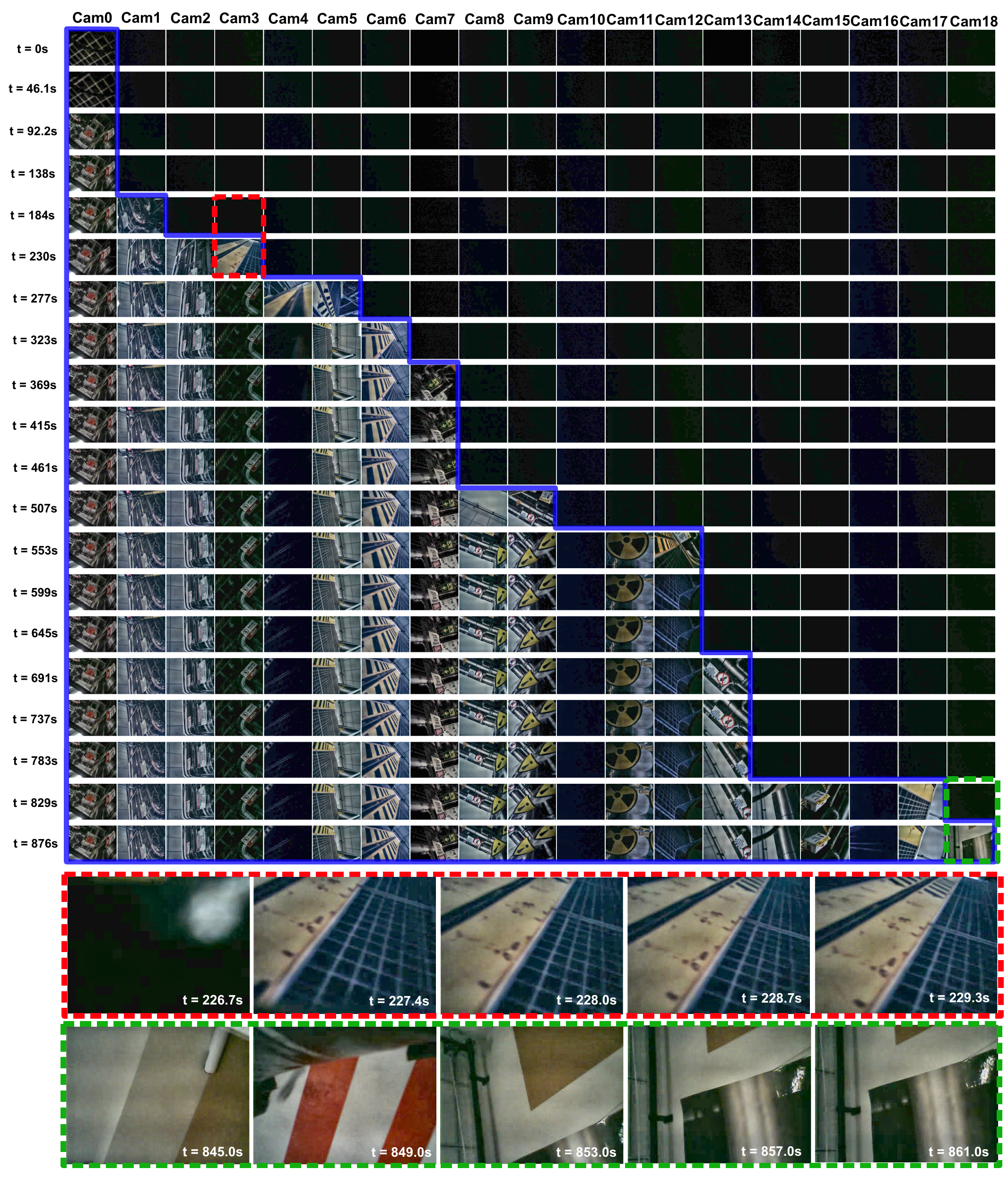}
  \caption{Camera-view visualization time series from 19 cameras across 20 time instances. Two example camera views during eversion are highlighted with red and green dashed outlines.}
  \label{fig:cameras}
\end{figure}

Fig.~\ref{fig:cameras} shows visualization of camera views on the complex course navigation task.

